\documentclass[sigconf]{acmart}
\usepackage{hyperref}

\usepackage{multirow} 
\usepackage{amsmath} 
\usepackage{bm}

\usepackage{algorithm}
\usepackage{algcompatible}

\usepackage{subfigure}
\usepackage{array}
\usepackage{longtable}
\usepackage{multirow}
\usepackage{bigstrut}
\usepackage{makecell}
\usepackage{enumitem}
\usepackage{booktabs}  
\usepackage{threeparttable}

\usepackage{xcolor,colortbl}
\usepackage{xurl,soul}
\usepackage{hhline}
\usepackage{gensymb}

\AtBeginDocument{%
  \providecommand\BibTeX{{%
    \normalfont B\kern-0.5em{\scshape i\kern-0.25em b}\kern-0.8em\TeX}}}

\copyrightyear{2023} 
\acmYear{2023} 
\setcopyright{rightsretained} 
\acmConference[SIGSPATIAL '23]{The 31st ACM International Conference on Advances in Geographic Information Systems}{November 13--16, 2023}{Hamburg, Germany}
\acmBooktitle{The 31st ACM International Conference on Advances in Geographic Information Systems (SIGSPATIAL '23), November 13--16, 2023, Hamburg, Germany}
\acmDOI{10.1145/3589132.3625596}
\acmISBN{979-8-4007-0168-9/23/11}

\begin{document}

\title{Devil in the Landscapes: Inferring Epidemic Exposure Risks from Street View Imagery}

\author{Zhenyu Han }
\email{hanzy19@tsinghua.edu.cn}
\orcid{0000-0001-9634-7962}
\affiliation{%
	\institution{Department of Electronic Engineering, BNRist, Tsinghua University}
	\city{Beijing}
	\country{China}
	\postcode{100084}
}

\author{Yanxin Xi}
\email{yanxin.xi@helsinki.fi}
\orcid{0000-0003-4715-2186}
\affiliation{%
	\institution{Department of Computer Science, University of Helsinki}
	\city{Helsinki}
	\country{Finland}
}

\author{Tong Xia}
\email{tx229@cam.ac.uk}
\orcid{0000-0002-6994-6318}
\affiliation{%
	\institution{Department of Computer Science and Technology, University of Cambridge}
	\city{Cambridge}
	\country{UK}
}

\author{Yu Liu}
\email{liuyu2419@126.com}
\affiliation{%
	\institution{Department of Electronic Engineering, BNRist, Tsinghua University}
	\city{Beijing}
	\country{China}
	\postcode{100084}
}

\author{Yong Li}
\authornote{Corresponding author.}
\email{liyong07@tsinghua.edu.cn}
\orcid{0000-0001-5617-1659}
\affiliation{%
	\institution{Department of Electronic Engineering, BNRist, Tsinghua University}
	\city{Beijing}
	\country{China}
	\postcode{100084}
}

%
%
%
%
%
%

\renewcommand{\shortauthors}{Han, et al.}

\begin{abstract}
Built environment supports all the daily activities and shapes our health. Leveraging informative street view imagery, previous research has established the profound correlation between the built environment and chronic, non-communicable diseases; however, predicting the exposure risk of infectious diseases remains largely unexplored. The person-to-person contacts and interactions contribute to the complexity of infectious disease, which is inherently different from non-communicable diseases. Besides, the complex relationships between street view imagery and epidemic exposure also hinder accurate predictions. To address these problems, we construct a regional mobility graph informed by the gravity model, based on which we propose a transmission-aware graph convolutional network (GCN) to capture disease transmission patterns arising from human mobility. Experiments show that the proposed model significantly outperforms baseline models by 8.54\% in weighted F1, shedding light on a low-cost, scalable approach to assess epidemic exposure risks from street view imagery.

\end{abstract}

\ccsdesc[500]{Applied computing~Life and medical sciences}
\ccsdesc[500]{Computing methodologies~Computer vision}
\ccsdesc[500]{Computing methodologies~Machine learning}

\keywords{Street View Imagery, Graph Convolutional Network, Computer Vision, Epidemic Risk, Ordinary Differential Equations, COVID-19}

\maketitle

\section{Introduction}\label{sec:Introduction}
\begin{figure}[t]
	\centering
	\includegraphics[width=0.35\textwidth]{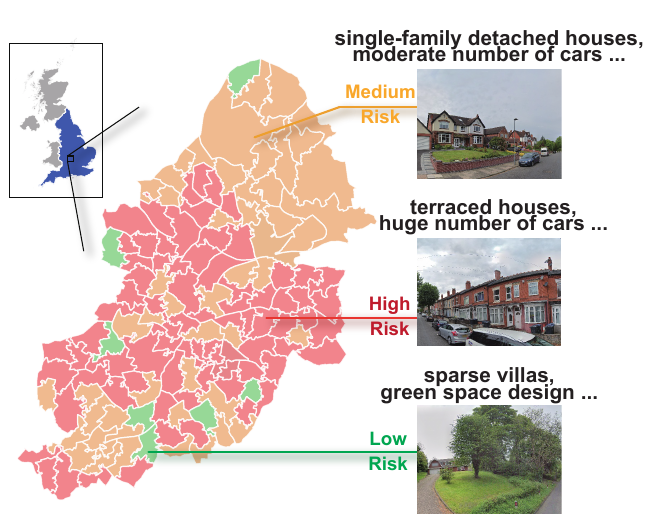}
	\caption{Illustration of epidemic exposure risk identification in Birmingham. In this study, we investigate the epidemic exposure risks for the whole England.}
	\label{fig:illustration}
	\vspace{-0.2in}
\end{figure}

With the rapid urbanization progress in the last century, more than 55\% of people live in cities surrounded by the built environment that provides the setting for all human activities, such as buildings, roads, green spaces, \emph{etc.}. The intimacy between human and urban built environment makes it a critical environmental determinant of health \cite{han2023healthy}, affecting both the physical and mental status of citizens. 

With the recent development of big data processing and deep learning technology, street view imagery provides a powerful data source to assess the built environment with rich information and high scalability \cite{li2022predicting}. In Figure \ref{fig:illustration}, we showcase that house types, the number of vehicles, the design of green spaces are possible transmission-related features for predicting epidemic exposure risks \cite{nguyen2021leveraging}. With publicly available street view imagery from map services or social media check-ins \cite{liu2023knowledge}, we can predict the epidemic exposure risk in most parts of the world, even in low- and middle-income countries that may lack detailed census data. 

However, accurately predicting epidemic exposure risk from street view imagery is challenging. First, infectious diseases are greatly influenced by the human mobility, which cannot be properly identified solely from street view imagery. Second, the distinct and complex transmission patterns arising from person-to-person contacts and interactions require different modeling approaches from traditional non-communicable diseases.

To overcome these challenges, we propose a novel model to identify epidemic exposure risks from street view imagery. Specifically, we construct a network of street view imagery to capture the regional transmission influence of infectious diseases, where we simulate the population flow inspired by Stouffer's law of population movement and Tobler's first law of geography. Based on the proposed network, we design a transmission-aware graph convolutional network (GCN) emulating epidemiological process inspired by Susceptible-Infectious-Recovered (SIR) model \cite{ross1916application}. Through the proposed model, we predict the global map of epidemic exposure risk in a low-cost, scalable manner, enlightening the design methodology for creating an epidemiologically resilient living environment through the power of geographic information system (GIS).

The contributions of this study can be summarized as follows:
\begin{itemize}
	\item We construct a regional transmission network informed by the street view imagery and human mobility, based on which the spatial correlation of epidemic can be accurately captured. 
	\item We propose a transmission-aware GCN model with epidemiological knowledge considering the distinct transmission patterns of infectious diseases.
	\item We conduct extensive experiments to validate the effectiveness of the proposed model, which outperforms the best baseline by $8.54\%$ in terms of weighted F1, $3.33\%$ in weighted precision, and $4.93\%$ in weighted recall.
\end{itemize}

\section{Preliminaries}\label{sec:Preliminaries}

The SIR model is a well-established epidemiological model that leverages the following ordinary differential equations (ODEs) to depict the dynamic of the epidemic:

\begin{align}
	\frac{\mathrm{d} \bm{S}(t)}{\mathrm{d} t} &=-\beta \frac{\bm{S}(t)\bm{I}(t)}{N},\\
	\frac{\mathrm{d} \bm{I}(t)}{\mathrm{d} t} &=\beta \frac{\bm{S}(t)\bm{I}(t)}{N} -\gamma \bm{I}(t), \\
	\frac{\mathrm{d} \bm{R}(t)}{\mathrm{d} t}&=\gamma \bm{I}(t).
\end{align}
The above model divides the whole population $N$ into four states: $\bm{S,I,R}$ for susceptible, infectious and recovered people accordingly. There are two learnable parameters: $\beta$ is the infection rate, while $\gamma$ is the recovery rate. It assumes a second-order contact between susceptible and infectious people for disease transmission as $\beta \bm{S}(t)\bm{I}(t)$, and a first-order natural recovery process $\gamma \bm{I}(t)$. Leveraging the calibrated SIR model, we can estimate the basic reproduction number $R_0$, which reflects the transmissibility of the target disease under specific urban scenarios as follows:
\begin{equation}
	R_0 = \beta / \gamma.
	\label{eq:r0}
\end{equation}
In this study, we use $R_0$ as an agent for epidemic exposure risk.

\section{Methods}
\begin{figure}[t]
	\centering
	\includegraphics[width=0.28\textwidth]{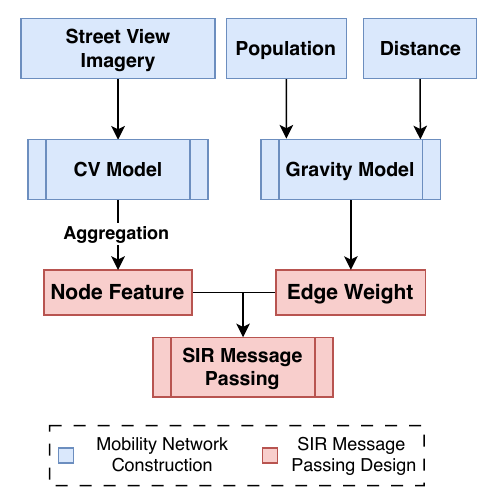}
	\caption{Framework overview.}
	\label{fig:framework}
	\vspace{-0.1in}
\end{figure}

We illustrate the framework of this study in Figure \ref{fig:framework}. We propose a transmission-aware GCN model, \emph{i.e.}, EpiGCN to infer the epidemic exposure risk through publicly available street view imagery. To capture the human mobility induced transmission of infectious diseases, we construct a regional mobility network, where the node feature represents geo-tagged imagery and the edge weight reflects the population flow simulated by the gravity model \cite{simini2012universal}. Inspired by the computational process of SIR ODEs, we design a novel message passing function that integrates the epidemiological model with representation learning. 

\subsection{Mobility Network Construction}\label{sec:gravity}
\begin{figure}[t]
	\centering
	\includegraphics[width=0.45\textwidth]{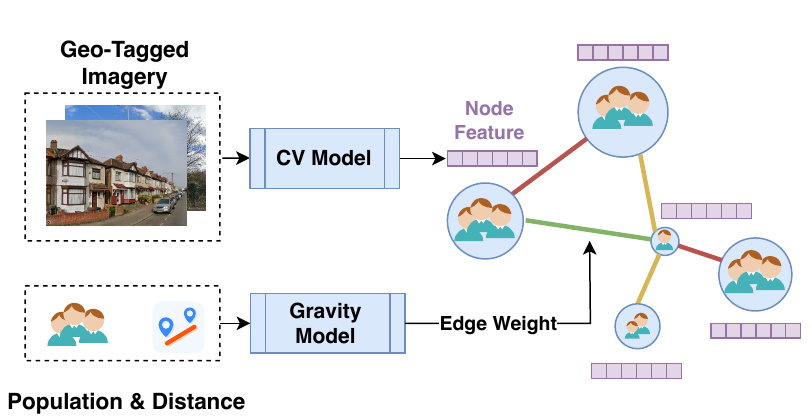}
	\caption{Illustration of the network construction process. The size of node represents the population, and the edge color represents the strength of population flow simulated by the gravity model.}
	\label{fig:graph_construct}
	\vspace{-0.1in}
\end{figure}

To capture the regional contact and transmission patterns invoked by human mobility, we first construct an MSOA level mobility network as shown in Figure \ref{fig:graph_construct}. Specifically, we have the graph $\mathcal{G} = (\mathcal{V}, \mathcal{E})$, where $\mathcal{V}$ is the set of MSOA and $\mathcal{E}$ is the set of mobility influences. 

Leveraging any CV backbone model $M_c$, we extract the MSOA level imagery feature by aggregating all the image dense embeddings of the corresponding MSOA, which serves as the node feature $\bm{h}_v, \forall v \in \mathcal{V}$. 

To depict the human mobility influence between MSOAs, we adopt the gravity model \cite{simini2012universal} to simulate regional population flows as the edge weight as follows:
\begin{equation}
	\label{eq:grav}
	e_{v,w} = \frac{N_v^\rho N_w^\theta}{\exp(d_{vw}/\delta)}, \forall v,w \in \mathcal{V},
\end{equation}
where $N_\star$ is the population number for MSOA $\star$, $d_{vw}$ is the Euclidean distance between $v,w$. We set the empirical parameters $\rho, \theta, \delta$ according to \cite{balcan2009multiscale}. 

Eq.\eqref{eq:grav} predicts the population flows as proportional to neighborhood population and inversely proportional to travel distance. The design of numerator is inspired by Stouffer's law of population movement that people are more attracted by regions with more social interaction opportunities. Besides, the denominator depicts Tobler's first law of geography that people tend to visit nearby places to reduce the travel cost.

\subsection{SIR Message Passing Design}\label{sec:mp}

Given the node feature $\bm{h}_v$ of target node $v$ and the neighborhood node set $\mathcal{N}_v$. First, we leverage three different linear layers to transform the node feature into susceptible (S), infectious (I), and recovered (R) embedding as follows:

\begin{equation*}
	\bm{S}_v = \sigma\Big(\bm{W}_S\cdot \bm{h}_v + \bm{b}_S\Big), \;\;\;\;
	\bm{I}_v = \sigma\Big(\bm{W}_I \cdot\bm{h}_v + \bm{b}_I\Big),\;\;\;\;
	\bm{R}_v = \sigma\Big(\bm{W}_R \cdot\bm{h}_v + \bm{b}_R\Big),
\end{equation*}
where $\bm{W}_{\star}$ is the weight matrix, $\bm{b}_{\star}$ is the bias matrix, and $\sigma$ is the rectified linear unit (ReLU) activation function.

The essence of SIR model lies in the second-order transmission process of $\beta\bm{S}(t)\bm{I}(t)$ and the first-order recovery process of $\gamma I(t)$. For the transmission process, we aggregate the I embeddings from neighborhood nodes according to the edge weights generated by the gravity model, which will be concatenated with local S embedding to depict the second-order interaction. It goes through a linear transformation to calculate the embedding of infections. For the recovery process, the only influence factor is the local I embedding. We use another linear transformation to depict the natural recovery process that happens in infectious people. The above process is shown in the following equations:
\begin{align}
	\bm{S}_v &= \bm{S}_v - \bm{W}_{tran} \cdot\mathrm{concat}\Big(\bm{S}_v, \sum_{w\in\mathcal{N}_v}e_{wv}\bm{I}_w \Big), \\
	\bm{I}_v &= \bm{I}_v + \bm{W}_{tran} \cdot\mathrm{concat}\Big(\bm{S}_v, \sum_{w\in\mathcal{N}_v}e_{wv}\bm{I}_w \Big) -\bm{W}_{recov}\bm{I}_v,   \\
	\bm{R}_v &= \bm{R}_v + \bm{W}_{recov} \cdot\bm{I}_v,
\end{align}
where $\bm{W}_{tran}\in\mathbb{R}^{2D\times D},\bm{W}_{recov}\in\mathbb{R}^{D\times D} $ represent the linear transformation for the transmission and recovery process accordingly, $D$ is the embedding size.

To get the epidemic exposure risk prediction, we further concatenate the local S, I, R embedding and use another linear transformation $\bm{W}_{output}\in\mathbb{R}^{3D\times D}$ to capture the epidemic exposure risk as follows:
\begin{equation}
	\hat{\bm{y}}^{main}_v = \mathrm{softmax}\Big(\bm{W}_{output} \cdot \mathrm{concat}(\bm{S}_v,\bm{I}_v,\bm{R}_v) \Big).
\end{equation}

\section{Experiments}\label{sec:Experiments}

\subsection{Setup of the Experiment}\label{sec:Datasets}
In this work, we study 6512 middle layer super output areas (MSOAs) in England, which are fine-grained census units with a mean population of 8236 and an average area of 19.5 $\mathrm{km}^2$. We collect the latest street view images from Google Map for each MSOA, where we uniformly sample 9 locations within the corresponding boundary \cite{guozhen}. As a result, we get 215,759 images for the whole of England MSOAs, where $76.8\%$ of them are captured after 2019. The street view images are available in $400\times300$, which will be randomly cropped into $224\times224$ before feeding into CV models. We download the MSOA level time series of COVID-19 cases in the second outbreak window (2020-09-01 to 2021-04-30) from the UK government. 

In this study, we adopt the basic reproduction number $R_0$ as an agent for epidemic exposure risk, which is a widely used metric to depict the severity of infectious diseases. Specifically, we calibrate the SIR model according to the COVID-19 time series, which provides a model-informed $R_0$ for each MSOA. Furthermore, we categorize the extracted $R_0$ in each MSOA according to the mean and standard deviation into three levels, which generates the epidemic exposure risk label $r\in \{0,1,2\}$ for low risk, medium risk, and high risk accordingly. The distribution of labels is demonstrated in Table \ref{tab:identified_epi_risk}. We randomly split the dataset into training, validation, and test sets in a $6:2:2$ ratio.

We implement the proposed EpiGCN in PyTorch, where we use ResNet18 initialized with ImageNet pre-trained weights as the CV backbone. Note that the whole architectures of the CV backbone are trainable. We adopt cross-entropy as the loss function. The implementation code of our model is available at \url{https://github.com/0oshowero0/EpidemicGCN}.

\begin{table}
	\caption{Distribution of epidemic exposure risk labels}
	\label{tab:identified_epi_risk}
	\begin{tabular}{c|ccc}
		\toprule
		Category &Low Risk&Medium Risk& High Risk\\
		Number & 1727 & 3820 & 1505\\
		Percentage & 27\%& 52\% & 23\%\\
		\bottomrule
	\end{tabular}
	\vspace{-0.1in}
\end{table}

\subsection{Baseline Models}\label{sec:baseline}
To the best of our knowledge, this is the first work to identify regional epidemic exposure risks through street view imagery. We adapt three commonly used paradigms in socioeconomic prediction task to validate the proposed method: feature based baselines (BOF \cite{wang2018urban}, SceneParse \cite{lee2021predicting}), end-to-end supervised CV baselines (ResNet18 \cite{resnet}, ViT-B/32 \cite{vit}), and unsupervised baselines (Urban2vec \cite{wang2020urban2vec}, READ \cite{han2020lightweight}, PG-SimCLR \cite{xi2022beyond}). Implementation details are summarized as below.

\begin{itemize}
	\item \textbf{BOF} \cite{wang2018urban}: Bag of feature method leverages HOG and GIST to extract geo-tagged imagery, which follows a random forest classifier to generate predictions. 
	\item \textbf{SceneParse} \cite{lee2021predicting}: SceneParse leverages the coverage ratio of each object to train an MLP for downstream tasks.
	\item \textbf{ResNet18} \cite{resnet}: An end-to-end deep learning CV model that follows pyramid architecture.
	\item \textbf{ViT-B/32} \cite{vit}: An end-to-end deep learning CV model that follows isotropic architecture.
	\item \textbf{Urban2vec} \cite{wang2020urban2vec}: An unsupervised model that constructs positive and negative image pairs according to the physical distance to guide CV model learning.
	\item \textbf{READ} \cite{han2020lightweight}: A semi-supervised model using a pretrained CV model that fine-tuned on downstream tasks using data pruning and dimensionality reduction technology.
	\item \textbf{PG-SimCLR} \cite{xi2022beyond}: An unsupervised model that use geographical distance and POI similarity to construct positive and negative image pairs and use attention module to fuse the embeddings.
	
\end{itemize}

We also implement two ablation models of the proposed EpiGCN:
\begin{itemize}
	\item \textbf{EpiGCN w/o gravity weight}: We delete the edge weight $e_{ij}$ generated by the gravity model as described in Sec.\ref{sec:gravity}.
	\item \textbf{EpiGCN w/o SIR message passing}: We replace the SIR passage passing design in Sec.\ref{sec:mp} by vanilla GCN.
\end{itemize}

\subsection{\textbf{Performance Analysis}}
The overall experiment results are reported in Table \ref{tab:main}. The baseline models are categorized into feature based baselines, end-to-end supervised CV baselines, and unsupervised CV baselines. From the results, we have the following observations and conclusions. 
\begin{itemize}
	\item Our proposed EpiGCN constantly outperforms the best baselines on all metrics statistically significantly, with $8.54\%$ higher weighted F1 than PG-SimCLR, $3.33\%$ higher in weighted precision than ViT-B/32, and $4.93\%$ higher in weighted recall than ResNet18.
	\item Compared with the two ablation models without gravity edge weight and SIR message passing, the full model outperforms by 65.9\% and 28.6\% in terms of weighted F1. This phenomenon demonstrates the effectiveness of the proposed improvements.
	\item In general, the performance of end-to-end supervised baselines surpasses that of unsupervised baselines, which in turn outperforms feature based baselines. Supervised baselines achieve the best performance for most metrics compared with other baselines. Most of the unsupervised baselines and feature based baselines perform poorly in terms of recall metrics, which is not convincing enough for mission-critical tasks such as epidemic exposure risk identification.
\end{itemize}

\begin{table}[t]
	\small
	\centering
	\caption{Performance comparison for epidemic exposure risk prediction. All the metrics are weighted ones. The average performance over 5 runs is reported. We bold the best performance, and ($\ast$) indicates p<0.01 significance over the best baseline metrics (underlined) in ANOVA test.}
	\label{tab:main}
	\begin{tabular}{m{4.2cm}<{\centering} |m{0.9cm}<{\centering} m{1cm}<{\centering} m{1cm}<{\centering}}
		\toprule
		Method & F1 & Precision & Recall \\ \midrule
		BOF \cite{wang2018urban}  &  0.4653   & 0.4711  & 0.4995 \\
		SenseParse \cite{lee2021predicting} & 0.4808  & 0.4883 &0.5085  \\ \hline
		ResNet18 \cite{resnet} & 0.4795   &0.4581 &$\underline{0.5360}$ \\
		ViT-B/32 \cite{vit} & 0.4907   & $\underline{0.5195}$ & 0.5329 \\ \hline
		Urban2vec \cite{wang2020urban2vec} & 0.4870   & 0.4950& 0.5190 \\
		READ \cite{han2020lightweight} &  0.4894  & 0.4976 & 0.5168\\
		PG-SimCLR \cite{xi2022beyond}&  $\underline{0.5014}$  &0.5100 & 0.5219\\ \hline
		EpiGCN w/o gravity edge weight & 0.3280   & 0.2836 & 0.4268 \\
		EpiGCN w/o SIR message passing &  0.4233  &0.3993  & 0.5371\\
		\textbf{EpiGCN (Ours)} & \textbf{0.5442}$\ast$   & \textbf{0.5368} $\ast$& \textbf{0.5624} $\ast$\\
		\bottomrule
	\end{tabular}
\end{table}

\section{Conclusion}\label{sec:Conclusion}
In this paper, we proposed a novel model that explicitly predicts regional epidemic exposure risks through street view imagery. Considering the inherently different transmission patterns of infectious diseases, we construct a network of street view imagery linked by regional population flow, based on which we propose a transmission-aware GCN model to capture the epidemic influence arising from human mobility. In the future, we will adopt transfer learning technology to further enhance the model performance in unseen regions, which enables a low-cost, scalable approach to assess how the built environment affects disease transmission.

\begin{acks}
This work was supported in part by The National Key Research and Development Program of China under grant 2022ZD0116402, the National Nature Science Foundation of China under U22B2057, 62171260, U1936217. 
\end{acks}

\bibliographystyle{ACM-Reference-Format}
\bibliography{references}


\begin{thebibliography}{15}


\ifx \showCODEN    \undefined \def \showCODEN     #1{\unskip}     \fi
\ifx \showDOI      \undefined \def \showDOI       #1{#1}\fi
\ifx \showISBNx    \undefined \def \showISBNx     #1{\unskip}     \fi
\ifx \showISBNxiii \undefined \def \showISBNxiii  #1{\unskip}     \fi
\ifx \showISSN     \undefined \def \showISSN      #1{\unskip}     \fi
\ifx \showLCCN     \undefined \def \showLCCN      #1{\unskip}     \fi
\ifx \shownote     \undefined \def \shownote      #1{#1}          \fi
\ifx \showarticletitle \undefined \def \showarticletitle #1{#1}   \fi
\ifx \showURL      \undefined \def \showURL       {\relax}        \fi
\providecommand\bibfield[2]{#2}
\providecommand\bibinfo[2]{#2}
\providecommand\natexlab[1]{#1}
\providecommand\showeprint[2][]{arXiv:#2}

\bibitem[Balcan et~al\mbox{.}(2009)]%
        {balcan2009multiscale}
\bibfield{author}{\bibinfo{person}{Duygu Balcan}, \bibinfo{person}{Vittoria
  Colizza}, \bibinfo{person}{Bruno Gon{\c{c}}alves}, \bibinfo{person}{Hao Hu},
  \bibinfo{person}{Jos{\'e}~J Ramasco}, {and} \bibinfo{person}{Alessandro
  Vespignani}.} \bibinfo{year}{2009}\natexlab{}.
\newblock \showarticletitle{Multiscale mobility networks and the spatial
  spreading of infectious diseases}.
\newblock \bibinfo{journal}{\emph{Proceedings of the National Academy of
  Sciences}} \bibinfo{volume}{106}, \bibinfo{number}{51}
  (\bibinfo{year}{2009}), \bibinfo{pages}{21484--21489}.
\newblock


\bibitem[Dosovitskiy et~al\mbox{.}(2020)]%
        {vit}
\bibfield{author}{\bibinfo{person}{Alexey Dosovitskiy}, \bibinfo{person}{Lucas
  Beyer}, \bibinfo{person}{Alexander Kolesnikov}, \bibinfo{person}{Dirk
  Weissenborn}, \bibinfo{person}{Xiaohua Zhai}, \bibinfo{person}{Thomas
  Unterthiner}, \bibinfo{person}{Mostafa Dehghani}, \bibinfo{person}{Matthias
  Minderer}, \bibinfo{person}{Georg Heigold}, \bibinfo{person}{Sylvain Gelly},
  {et~al\mbox{.}}} \bibinfo{year}{2020}\natexlab{}.
\newblock \showarticletitle{An image is worth 16x16 words: Transformers for
  image recognition at scale}.
\newblock \bibinfo{journal}{\emph{arXiv preprint arXiv:2010.11929}}
  (\bibinfo{year}{2020}).
\newblock


\bibitem[Han et~al\mbox{.}(2020)]%
        {han2020lightweight}
\bibfield{author}{\bibinfo{person}{Sungwon Han}, \bibinfo{person}{Donghyun
  Ahn}, \bibinfo{person}{Hyunji Cha}, \bibinfo{person}{Jeasurk Yang},
  \bibinfo{person}{Sungwon Park}, {and} \bibinfo{person}{Meeyoung Cha}.}
  \bibinfo{year}{2020}\natexlab{}.
\newblock \showarticletitle{Lightweight and robust representation of economic
  scales from satellite imagery}. In \bibinfo{booktitle}{\emph{Proceedings of
  the AAAI Conference on Artificial Intelligence}}, Vol.~\bibinfo{volume}{34}.
  \bibinfo{pages}{428--436}.
\newblock


\bibitem[Han et~al\mbox{.}(2023)]%
        {han2023healthy}
\bibfield{author}{\bibinfo{person}{Zhenyu Han}, \bibinfo{person}{Tong Xia},
  \bibinfo{person}{Yanxin Xi}, {and} \bibinfo{person}{Yong Li}.}
  \bibinfo{year}{2023}\natexlab{}.
\newblock \showarticletitle{Healthy Cities, A comprehensive dataset for
  environmental determinants of health in England cities}.
\newblock \bibinfo{journal}{\emph{Scientific Data}} \bibinfo{volume}{10},
  \bibinfo{number}{1} (\bibinfo{year}{2023}), \bibinfo{pages}{165}.
\newblock
\urldef\tempurl%
\url{https://doi.org/10.1038/s41597-023-02060-y}
\showDOI{\tempurl}


\bibitem[He et~al\mbox{.}(2016)]%
        {resnet}
\bibfield{author}{\bibinfo{person}{Kaiming He}, \bibinfo{person}{Xiangyu
  Zhang}, \bibinfo{person}{Shaoqing Ren}, {and} \bibinfo{person}{Jian Sun}.}
  \bibinfo{year}{2016}\natexlab{}.
\newblock \showarticletitle{Deep residual learning for image recognition}. In
  \bibinfo{booktitle}{\emph{Proceedings of the {IEEE/CVF} Conference on
  Computer Vision and Pattern Recognition {(CVPR)}}}.
  \bibinfo{pages}{770--778}.
\newblock


\bibitem[Lee et~al\mbox{.}(2021)]%
        {lee2021predicting}
\bibfield{author}{\bibinfo{person}{Jihyeon Lee}, \bibinfo{person}{Dylan Grosz},
  \bibinfo{person}{Burak Uzkent}, \bibinfo{person}{Sicheng Zeng},
  \bibinfo{person}{Marshall Burke}, \bibinfo{person}{David Lobell}, {and}
  \bibinfo{person}{Stefano Ermon}.} \bibinfo{year}{2021}\natexlab{}.
\newblock \showarticletitle{Predicting Livelihood Indicators from
  Community-Generated Street-Level Imagery}. In
  \bibinfo{booktitle}{\emph{Proceedings of the AAAI Conference on Artificial
  Intelligence}}, Vol.~\bibinfo{volume}{35}. \bibinfo{pages}{268--276}.
\newblock


\bibitem[Li et~al\mbox{.}(2022)]%
        {li2022predicting}
\bibfield{author}{\bibinfo{person}{Tong Li}, \bibinfo{person}{Shiduo Xin},
  \bibinfo{person}{Yanxin Xi}, \bibinfo{person}{Sasu Tarkoma},
  \bibinfo{person}{Pan Hui}, {and} \bibinfo{person}{Yong Li}.}
  \bibinfo{year}{2022}\natexlab{}.
\newblock \showarticletitle{Predicting Multi-level Socioeconomic Indicators
  from Structural Urban Imagery}. In \bibinfo{booktitle}{\emph{Proceedings of
  the 31st ACM International Conference on Information \& Knowledge
  Management}}. \bibinfo{pages}{3282--3291}.
\newblock


\bibitem[Liu et~al\mbox{.}(2023)]%
        {liu2023knowledge}
\bibfield{author}{\bibinfo{person}{Yu Liu}, \bibinfo{person}{Xin Zhang},
  \bibinfo{person}{Jingtao Ding}, \bibinfo{person}{Yanxin Xi}, {and}
  \bibinfo{person}{Yong Li}.} \bibinfo{year}{2023}\natexlab{}.
\newblock \showarticletitle{Knowledge-infused contrastive learning for urban
  imagery-based socioeconomic prediction}. In
  \bibinfo{booktitle}{\emph{Proceedings of the ACM Web Conference 2023}}.
  \bibinfo{pages}{4150--4160}.
\newblock


\bibitem[Nguyen et~al\mbox{.}(2021)]%
        {nguyen2021leveraging}
\bibfield{author}{\bibinfo{person}{Quynh~C Nguyen}, \bibinfo{person}{Jessica~M
  Keralis}, \bibinfo{person}{Pallavi Dwivedi}, \bibinfo{person}{Amanda~E Ng},
  \bibinfo{person}{Mehran Javanmardi}, \bibinfo{person}{Sahil Khanna},
  \bibinfo{person}{Yuru Huang}, \bibinfo{person}{Kimberly~D Brunisholz},
  \bibinfo{person}{Abhinav Kumar}, {and} \bibinfo{person}{Tolga Tasdizen}.}
  \bibinfo{year}{2021}\natexlab{}.
\newblock \showarticletitle{Leveraging 31 million Google street view images to
  characterize built environments and examine County health outcomes}.
\newblock \bibinfo{journal}{\emph{Public Health Reports}}
  \bibinfo{volume}{136}, \bibinfo{number}{2} (\bibinfo{year}{2021}),
  \bibinfo{pages}{201--211}.
\newblock


\bibitem[Ross(1916)]%
        {ross1916application}
\bibfield{author}{\bibinfo{person}{Ronald Ross}.}
  \bibinfo{year}{1916}\natexlab{}.
\newblock \showarticletitle{An application of the theory of probabilities to
  the study of a priori pathometry.—Part I}.
\newblock \bibinfo{journal}{\emph{Proceedings of the Royal Society of London.
  Series A, Containing papers of a mathematical and physical character}}
  \bibinfo{volume}{92}, \bibinfo{number}{638} (\bibinfo{year}{1916}),
  \bibinfo{pages}{204--230}.
\newblock


\bibitem[Simini et~al\mbox{.}(2012)]%
        {simini2012universal}
\bibfield{author}{\bibinfo{person}{Filippo Simini}, \bibinfo{person}{Marta~C
  Gonz{\'a}lez}, \bibinfo{person}{Amos Maritan}, {and}
  \bibinfo{person}{Albert-L{\'a}szl{\'o} Barab{\'a}si}.}
  \bibinfo{year}{2012}\natexlab{}.
\newblock \showarticletitle{A universal model for mobility and migration
  patterns}.
\newblock \bibinfo{journal}{\emph{Nature}} \bibinfo{volume}{484},
  \bibinfo{number}{7392} (\bibinfo{year}{2012}), \bibinfo{pages}{96--100}.
\newblock


\bibitem[Wang et~al\mbox{.}(2018)]%
        {wang2018urban}
\bibfield{author}{\bibinfo{person}{Wenshan Wang}, \bibinfo{person}{Su Yang},
  \bibinfo{person}{Zhiyuan He}, \bibinfo{person}{Minjie Wang},
  \bibinfo{person}{Jiulong Zhang}, {and} \bibinfo{person}{Weishan Zhang}.}
  \bibinfo{year}{2018}\natexlab{}.
\newblock \showarticletitle{Urban perception of commercial activeness from
  satellite images and streetscapes}. In \bibinfo{booktitle}{\emph{Companion
  Proceedings of the The Web Conference 2018}}. \bibinfo{pages}{647--654}.
\newblock


\bibitem[Wang et~al\mbox{.}(2020)]%
        {wang2020urban2vec}
\bibfield{author}{\bibinfo{person}{Zhecheng Wang}, \bibinfo{person}{Haoyuan
  Li}, {and} \bibinfo{person}{Ram Rajagopal}.} \bibinfo{year}{2020}\natexlab{}.
\newblock \showarticletitle{Urban2vec: Incorporating street view imagery and
  pois for multi-modal urban neighborhood embedding}. In
  \bibinfo{booktitle}{\emph{Proceedings of the AAAI Conference on Artificial
  Intelligence}}, Vol.~\bibinfo{volume}{34}. \bibinfo{pages}{1013--1020}.
\newblock


\bibitem[Xi et~al\mbox{.}(2022)]%
        {xi2022beyond}
\bibfield{author}{\bibinfo{person}{Yanxin Xi}, \bibinfo{person}{Tong Li},
  \bibinfo{person}{Huandong Wang}, \bibinfo{person}{Yong Li},
  \bibinfo{person}{Sasu Tarkoma}, {and} \bibinfo{person}{Pan Hui}.}
  \bibinfo{year}{2022}\natexlab{}.
\newblock \showarticletitle{Beyond the First Law of Geography: Learning
  Representations of Satellite Imagery by Leveraging Point-of-Interests}. In
  \bibinfo{booktitle}{\emph{Proceedings of the ACM Web Conference 2022}}.
  \bibinfo{pages}{3308--3316}.
\newblock


\bibitem[Zhang et~al\mbox{.}(2023)]%
        {guozhen}
\bibfield{author}{\bibinfo{person}{Guozhen Zhang}, \bibinfo{person}{Jinhui Yi},
  \bibinfo{person}{Jian Yuan}, \bibinfo{person}{Yong Li}, {and}
  \bibinfo{person}{Depeng Jin}.} \bibinfo{year}{2023}\natexlab{}.
\newblock \showarticletitle{DAS: Efficient Street View Image Sampling for Urban
  Prediction}.
\newblock \bibinfo{journal}{\emph{ACM Transactions on Intelligent Systems and
  Technology}} \bibinfo{volume}{14}, \bibinfo{number}{2}
  (\bibinfo{year}{2023}), \bibinfo{pages}{1--20}.
\newblock


\end{thebibliography}


\end{document}